\documentclass[10pt,twocolumn,letterpaper]{article}

\usepackage[pagenumbers]{iccv} 
\usepackage{multicol}
\usepackage{multirow}
\usepackage[accsupp]{axessibility}  
\definecolor{iccvblue}{rgb}{0.21,0.49,0.74}
\usepackage[pagebackref,breaklinks,colorlinks,allcolors=iccvblue]{hyperref}

\def\paperID{11} 
\def\confName{ICCV}
\def\confYear{2025}

\title{PTQAT: A Hybrid Parameter-Efficient Quantization Algorithm \\ for 3D Perception Tasks}

\author{
Xinhao Wang\quad
~Zhiwei Lin\quad
~Zhongyu Xia\quad
~ Yongtao Wang\thanks{Corresponding author.} \\
{Wangxuan Institute of Computer Technology, Peking University} \\ 
{\tt\small \{wangxinhao, zwlin, xiazhongyu, wyt\}@pku.edu.cn} \\
}

\begin{document}
\maketitle
\begin{abstract}
Post-Training Quantization (PTQ) and Quantization-Aware Training (QAT) represent two mainstream model quantization approaches. 
However, PTQ often leads to unacceptable performance degradation in quantized models, while QAT imposes substantial GPU memory requirements and extended training time due to weight fine-tuning.
%
In this paper, we propose PTQAT, a novel general hybrid quantization algorithm for the efficient deployment of 3D perception networks. 
%
To address the speed-accuracy trade-off between PTQ and QAT, our method selects critical layers for QAT fine-tuning and performs PTQ on the remaining layers. 
%
%
Contrary to intuition, fine-tuning the layers with smaller output discrepancies before and after quantization, rather than those with larger discrepancies, actually leads to greater improvements in the model’s quantization accuracy. This means we better compensate for quantization errors during their propagation, rather than addressing them at the point where they occur.
%
The proposed PTQAT achieves similar performance to QAT with more efficiency by freezing nearly 50\% of quantifiable layers.
Additionally, PTQAT is a universal quantization method that supports various quantization bit widths (4 bits) as well as different model architectures, including CNNs and Transformers. 
%
The experimental results on nuScenes across diverse 3D perception tasks, including object detection, semantic segmentation, and occupancy prediction, show that our method consistently outperforms QAT-only baselines.
Notably, it achieves 0.2\%-0.9\% NDS and 0.3\%-1.0\% mAP gains in object detection, 0.3\%-2.0\% mIoU gains in semantic segmentation and occupancy prediction while fine-tuning fewer weights.
\end{abstract}    
\section{Introduction}
\label{sec:intro}

\begin{figure*}
  \centering
   \includegraphics[width=\linewidth]{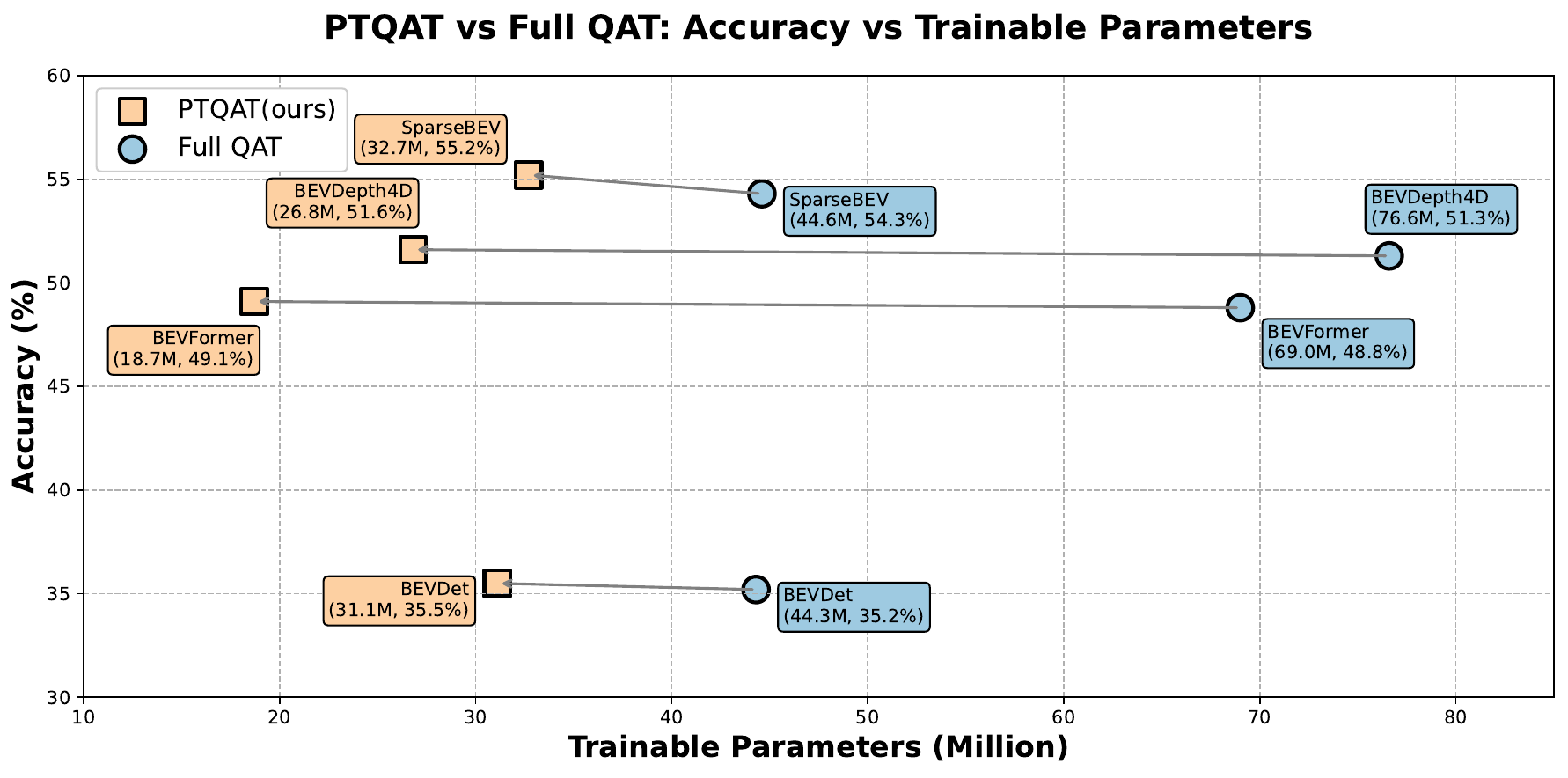}
   \caption{Comparison (NuScenes Dataset Score and trainable parameter numbers) results for QAT only method and the proposed PTQAT method. 
   All models are quantized to W4(4-bit weight) .
   \textbf{Compared to standard QAT, our approach trains fewer parameter numbers, achieving higher NDS with lower training costs.}}
   \label{fig:precision}
\end{figure*}


Neural networks are advancing rapidly in the fields of Computer Vision, especially in 3D perception tasks. 
3D perception tasks, including 3D object detection, BEV (Bird-Eye-View) semantic segmentation, and occupancy prediction, involve understanding and interpreting the 3D structure of the environment from sensory data, enabling machines to perceive depth, shapes, and spatial relationships like humans. 
A major application of 3D perception is autonomous driving, which demands exceptionally high accuracy and real-time model inference due to its critical impact on safety.
However, to achieve higher accuracy, researchers often increase the number of model parameters or design more complex model architectures, leading to unacceptable inference latency and GPU memory usage for 3D perception models on edge devices.
Therefore, it is crucial to lightweight the 3D perception model for real-world usage and deployment.
%
%
%


Recently, model quantization has become a crucial method in model light-weighting.
Model quantization can be divided into two main categories, \textit{i.e.}, post-training quantization (PTQ) and quantization-aware training (QAT). PTQ converts a pre-trained neural network to lower precision after training, enabling efficient deployment with minimal accuracy loss. QAT simulates quantization effects during training, allowing models to adapt to reduced precision and maintain higher accuracy.
%
In practical applications, PTQ is thousands of times faster than QAT. 
However, QAT is approximately 10\% more accurate than PTQ~\cite{fu2025quantization}.
Therefore, finding a trade-off between the two methods is important.


In this paper, we propose a general hybrid quantization method, PTQAT, to combine the benefits of PTQ and QAT. 
Our core idea is to utilize QAT to fine-tune the partial weights of models, compensating for the impact of PTQ on accuracy. 
Specifically, in extremely low-bit (\textit{e.g.}, 4 bits) quantization, the traditional PTQ methods cannot accurately quantize weights that are unevenly distributed or have extreme values, resulting in performance degradation in the quantized model. 
To address this issue, we aim to identify the layer with a significant positioning error in units of layers and adopt QAT to fine-tune its weights. 
An intuitive idea is to compare the difference in the layer's output before and after quantization, and quantize those layers with a large difference. However, the opposite way can actually improve quantization results. This phenomenon reveals a fundamental principle: \textbf{Quantization errors matter most where they propagate, not where they appear}. Adjusting the method to track the flow of errors, rather than focusing solely on where the errors originate, can lead to further improvements. The details of analysis are presented in Section~\ref{subsection:method-qat}.
After identifying layers to fine-tune, we freeze all other parameters and perform QAT on these layers to control the quantization error within a reasonable range.

The proposed PTQAT is a universal quantization method that can be applied to various 3D perception models. enabling fast quantization with minimal loss of accuracy. As shown in Figure~\ref{fig:precision}, we can see that our method can match or even surpass the performance of full-parameter QAT methods while requiring fewer parameter numbers to be fine-tuned, in some cases, only one-quarter as many.
It is worth mentioning that our quantization method primarily focuses on the real-world deployment of the model without any hardware customization and optimization, allowing the quantized model to be deployed through tools such as NVIDIA TensorRT. Our quantization strategy is a simple uniform symmetric quantization, and we have successfully deployed an INT8 quantized 3D perception model to a TensorRT engine . The results are listed in Section~\ref{subsection:tensorrt}.

Our contributions are listed as follows: 
\begin{itemize}
    \item We introduce a general hybrid quantization approach, PTQAT, designed to leverage the advantages of both PTQ and QAT.    
    \item We investigate approaches for reducing the number of weights requiring fine-tuning during QAT and conclude that mitigating quantization errors should target not the layers where the errors originate, but rather the process of error propagation itself. To the best of our knowledge, no previous quantization methods have explored this direction.
    \item Experimental results on nuScenes show that PTQAT achieves favorable performance on various 3D perception tasks, including 3D object detection, BEV segmentation, and occupancy prediction. Moreover, we utilize uniform symmetric quantization, successfully deploying an INT8 quantized model to a TensorRT engine. This demonstrates the versatility and practicality of our method in real-world deployment scenarios.

    
\end{itemize}
\section{Related Work}
\label{sec:related}

\subsection{3D Perception Tasks}


3D perception tasks include 3D object detection~\cite{BEVDet, BEVDepth, BEVStereo, HENet, DETR3D, PETR, StreamPetr, SparseBEV, RoPETR, focalformer3d, Real-Aug, LargeKernel3D, RCBEVDet, Bevfusion, SimpleBEV, Ea-lss}, BEV segmentation~\cite{Bevsegformer, CoBEVT, zhao2024improving, kong2023rethinking, peng2024oa, hu2021bidirectional}, occupancy prediction~\cite{roldao2020lmscnet, cheng2021s3cnet, Pointocc, Panossc, GEOcc, Octreeocc, Co-occ, sze2024real, OccFusion}, etc. 

3D object detection is widely regarded as the principal task. 
%
For camera-based 3D object detection, Lift-splat-shoot (LSS)~\cite{LSS} 
estimates depth distributions to project 2D image features into 3D space, laying the groundwork for subsequent advances. 
Numerous methods, such as BEVDet~\cite{BEVDet}, BEVDepth~\cite{BEVDepth}, BEVStereo~\cite{BEVStereo} and HENet~\cite{HENet}, have further refined this paradigm.
%
An alternative query-based paradigm has been proposed by DETR3D~\cite{DETR3D} and PETR~\cite{PETR}, wherein detection is formulated using predefined 3D anchor or query points;
%
%
Notable works building upon this paradigm also include StreamPetr~\cite{StreamPetr}, SparseBEV~\cite{SparseBEV}, and RoPETR~\cite{RoPETR}. 
Multi-tasking methods, exemplified by BEVFormer~\cite{Bevformer} and PETRv2~\cite{Petrv2}, simultaneously address 3D object detection and BEV semantic segmentation. 
%
Beyond cameras, autonomous vehicles commonly integrate additional sensors, such as LiDAR and radar. 
%
Methods including FocalFormer3D~\cite{focalformer3d}, Real-Aug~\cite{Real-Aug}, and LargeKernel3D~\cite{LargeKernel3D} are designed to extract features from point cloud data to facilitate 3D perception. 
Furthermore, RCBEVDet~\cite{RCBEVDet}, BEVFusion~\cite{Bevfusion}, SimpleBEV~\cite{SimpleBEV}, and EA-LSS~\cite{Ea-lss} are multi-modal methods that combine multiple sensors to predict. 

BEV semantic segmentation models classify the semantic category of each pixel in a vehicle's bird's-eye view (BEV) map.
For instance, BEVSegFormer~\cite{Bevsegformer} enhances camera features with a deformable transformer encoder and then employs a transformer decoder to generate segmentation results. 
CoBEVT~\cite{CoBEVT} leverages vehicle-to-vehicle communication to fuse camera features from multiple views and agents, enabling collaborative BEV map prediction. 
%
Other methods~\cite{zhao2024improving} decompose the segmentation task into two stages: BEV map reconstruction and RGB-to-BEV feature alignment. 
Alternatively, models like RangeFormer~\cite{kong2023rethinking}, Oa-cnns~\cite{peng2024oa}, and BPNet~\cite{hu2021bidirectional} transform 3D data (\textit{e.g.}, point clouds or voxels) into a 2D representation to perform semantic segmentation with established 2D convolution neural networks (CNNs).  
Furthermore, RCBEVDet++~\cite{RCBEVDet++} and Bevcar~\cite{Bevcar}  fuse camera and radar point cloud features to facilitate multi-task prediction.
 
Occupancy prediction methods are typically categorized as LiDAR-based, camera-based, or multi-modal.
%
LiDAR-based models like LMS-Net~\cite {roldao2020lmscnet}, S3C-Net~\cite {cheng2021s3cnet}, and PointOcc~\cite{Pointocc} employ an encoder-decoder architecture to process point cloud features and subsequently infer the scene's complete and dense occupancy.
Camera-based methods~\cite{Panossc, GEOcc, Octreeocc} first extract feature maps from multiple camera images, then transform them from 2D to 3D space, and finally fuse spatial and temporal information.
Multi-modal methods~\cite{sze2024real} leverage the strengths of various sensors to mitigate the limitations of single-modality approaches. 
For instance, models such as Co-Occ~\cite{Co-occ}, TEOcc~\cite{TEOcc}, and OccFusion~\cite{OccFusion} project 2D image features into 3D space and fuse them with point cloud features for improved spatial understanding.

\subsection{Post-Training Quantization}

Model quantization is a highly effective technique for model compression.
Core configuration choices in quantization include selecting the data to be quantized (weights and activations), choosing between symmetric and asymmetric schemes, and determining the bit-width.
Among these, uniform symmetric quantization is one of the most widely adopted methods.
The process for $b$ bits quantization is defined as: 
\begin{equation}
x^{\mathbb{Z}}=\operatorname{clip}\left(\left\lfloor\frac{x}{s}\right\rceil, -2^{b-1},2^{b-1}-1\right),
\label{eq:quant}
\end{equation}
where $\lfloor \cdot \rceil$ denotes the rounding function, $clip(\cdot, a, b)$ clips a value to the range $[a, b]$, and $s \in \mathbb{R}^+$ is the scaling factor, which is calculated as follows:
\begin{equation}
s = \cfrac{max(abs(max(x)), abs(min(x)))}{2^{b-1}-1}.
\label{eq:cal-s}
\end{equation}
However, naively applying this approach can lead to significant performance degradation.

To mitigate quantization errors in CNN models, various methods have been developed. For instance, some approaches~\cite{wang2020towards} employ bit splitting to reduce channel-wise differences in activation values. 
BRECQ~\cite{Brecq} utilizes block reconstruction for quantization, guided by the Fisher Information Matrix, to optimize the quantization process.
AdaRound~\cite{adaround} introduces an improved weight rounding mechanism for post-training quantization that adapts based on data and task-specific loss.
To account for activation distributions, QDrop~\cite{Qdrop} randomly enables or disables activation quantization during each forward pass.
Similar to our work, PD-Quant~\cite{Pd-quant} determines the scaling factor by comparing the pairwise distance loss of each layer before and after quantization.

With the success of the Transformer architecture, research has increasingly focused on quantizing these models.
Previous work~\cite{liu2021post} focuses on quantizing all weights and inputs involved in the matrix multiplication. 
FQ-ViT~\cite{lin2021fq} proposes a Power-of-Two Factor method to address significant inter-channel variations in LayerNorm inputs, and introduced Log-Int-Softmax to preserve the non-uniform distribution of attention maps. 
PTQ4ViT~\cite{Ptq4vit} employs a dual uniform quantization method to reduce errors in activations following softmax and GELU functions. 
By decoupling the quantization and inference processes, RepQ-ViT~\cite{Repq-vit} uses scale re-parameterization to generate simple, hardware-friendly quantizers.
Finally, AdaLog~\cite{Adalog} proposes a novel non-uniform quantizer that optimizes the logarithmic base to accommodate power-law distributed activations while remaining hardware-friendly.

\subsection{Quantization-Aware Training}

Quantization-Aware Training (QAT) is another prominent quantization technique.
In contrast to post-training methods, QAT integrates the quantization process into the network training loop, co-optimizing quantization parameters and model weights to minimize task loss on a labeled dataset.

Several QAT methods have been proposed. LSQ\cite{esser2019learned}, for example, treats the quantization scale as a trainable parameter, whereas traditional approaches often quantize only the weights
EWGS~\cite{lee2021network} improves the Straight-Through Estimator (STE), a common gradient approximation for quantization, by introducing an element-wise gradient scaling algorithm that adaptively adjusts each gradient element.

For Transformer-based models, Q-ViT~\cite{Q-vit} employs an Information Rectification Module (IRM) and a Distribution-Guided Distillation (DGD) scheme to mitigate information distortion in low-bit self-attention maps.
Q-BERT~\cite{Q-bert} leverages the Hessian spectrum to assess tensor sensitivity for mixed-precision quantization, achieving 3-bit weight and 8-bit activation quantization.
QAT principles have also been extended to Binarized Neural Networks (BNNs).
For example, ReActNet~\cite{Reactnet} constructs a BNN by modifying and binarizing a compact real-valued network and then uses a distribution loss function to train the binary network to mimic the output distribution of its full-precision counterpart.

\section{Method}


This section presents the implementation details of our proposed method, PTQAT. Our approach demonstrates that applying Quantization-Aware Training to only a subset of a model's weights can achieve performance comparable or superior to that of fine-tuning the entire network.
%

\begin{figure}[t]
  \centering
   \includegraphics[width=\linewidth]{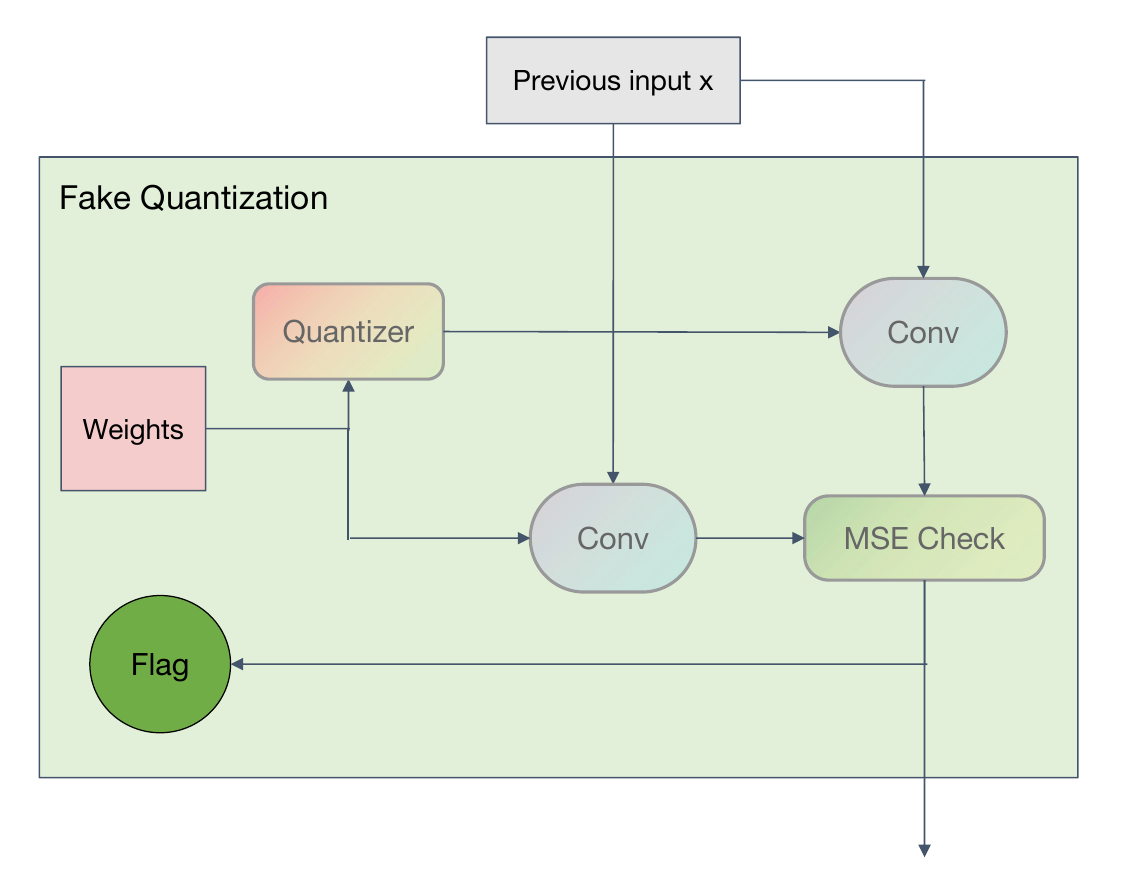}

   \caption{Illustration of \textbf{PTQ Pre-Check} in one Conv2d layer. PTQ Pre-Check performs a convolution using both quantized and non-quantized weights, then evaluates the impact of quantization on the layer by computing the mean squared error (MSE) between the resulting outputs, which is used to determine whether or not to quantize that layer.}
   
   \label{fig:precheck}
\end{figure}

\subsection{PTQ Pre-Check}
Fine-tuning a large number of weights during Quantization-Aware Training (QAT) can impede convergence. As the number of tunable weights $N$ increases, the dimensionality of the optimization space grows, proportional to $O(N)$. This expansion results in a more complex loss landscape populated with numerous local minima, increasing the risk of the model converging to a suboptimal solution. Moreover, reducing the number of fine-tuned weights intuitively improves the training efficiency of QAT. Consequently, the crucial challenge is to determine which layers require fine-tuning. 

To address this challenge, we introduce a PTQ Pre-Check procedure, as illustrated in~\ref{fig:precheck}.
This procedure begins by quantizing the entire model using Post-Training Quantization (PTQ) to gather the outputs of each layer before and after this process.
Specifically, we apply uniform symmetric quantization to  the trained model weights, as shown in Eq.\ref{eq:quant}. The original layer output $\mathbb{X}$ and the post-quantization output $\mathbb{X}_Q$ are defined as:

\begin{equation}
\begin{aligned}
& \mathbb{X} = F(\mathbb{W}, A), \\
& \mathbb{X}_Q = F(Q(\mathbb{W}), A) ,
\label{eq:output}
\end{aligned}
\end{equation}
where $F$ represents the layer's operation, $W$ and $A$ are the full-precision weights and input activations, respectively, and $Q$ denotes the uniform symmetric quantization function.
After obtaining both outputs, we calculate the distance between them using the Mean Squared Error (MSE) to quantify the quantization error:
\begin{equation}
\begin{aligned}
& \text{MSE}=\frac{1}{n} \sum_{i=1}^{n}\left(y_{i}-\hat{y}_{i}\right)^{2}, \\
& Dis(\mathbb{X}_Q, \mathbb{X}) = \text{MSE}(\mathbb{X}_Q, \mathbb{X}). 
\label{eq:dis}
\end{aligned}
\end{equation}

\subsection{QAT Fine-Tune}
\label{subsection:method-qat}
Empirically, a significant difference between a layer's output before and after quantization indicates that PTQ has introduced substantial error, suggesting this layer as a candidate for fine-tuning.
\begin{figure*}
  \centering
   \includegraphics[width=0.9\linewidth]{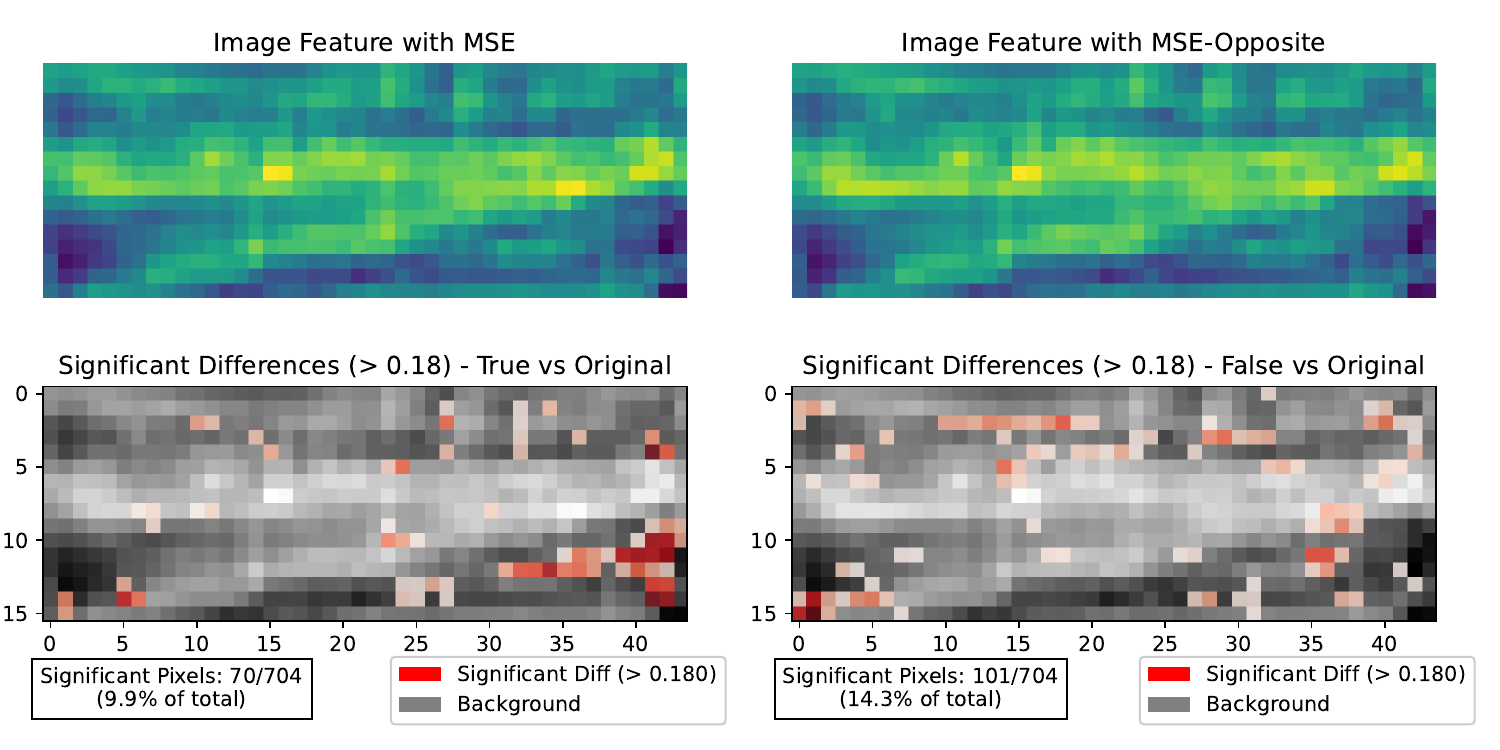}
   \caption{Euclidean distance visualization of image features extracted from ResNet50. Image feature with MSE comes from a model quantized by our PTQAT method, image feature with MSE-Opposite comes from a model quantized by fine-tuning layers with a larger output difference, which is opposite to our method. 
   The bottom results show the Euclidean distance between each quantized feature and the original feature. 
   Our method generates features with a 4.4\% less significant difference.}
   \label{fig:heatmap}
\end{figure*}
However, our findings present a counter-intuitive perspective. Layers with minimal initial quantization error are not necessarily robust. Their low error is observed only when receiving full-precision inputs. During QAT, the inputs to these layers undergo significant distributional shifts caused by the fine-tuning of upstream weights. Because the weights of these low-error layers are not being adjusted,they become highly vulnerable to these input variations. Our strategy, therefore, involves fine-tuning the layers that are susceptible to these propagated errors, rather than the layers where the errors originate. This allows the network to learn to compensate for distortions during the optimization process, effectively managing error propagation instead of merely correcting it at the source.

To validate this approach, we visualize the feature maps produced by different quantization strategies and calculate their Euclidean distance from the full-precision baseline, as shown in Figure ~\ref{fig:heatmap}. We the conventional approach—fine-tuning layers with the largest initial output difference—may reduce the magnitude of initial errors but often causes these errors to diffuse across the network,leading to a higher overall proportion of outliers. In contrast, our method, by focusing on error propagation, effectively contains the errors within their source layers. Although some large errors might remain at the source, this containment strategy prevents widespread distortion, resulting in a lower proportion of outlier errors and ultimately achieving superior quantization performance.

We determine whether a layer's weights require fine-tuning based on the calculated distance $Dis(\mathbb{X}_Q, \mathbb{X})$ and a predefined hyperparameter $\theta$:

\begin{equation}
\text{Fine-tune} = 
\begin{cases} 
    \text{True}, & \text{if } Dis(\mathbb{X}_Q, \mathbb{X}) < \theta \\
    \text{False}, & \text{otherwise}
\end{cases}.
\label{eq:flag}
\end{equation}
Once the layers for fine-tuning are identified, we freeze the weights of the other layers and apply QAT exclusively to this selected subset.

A key challenge during back-propagation is handling the non-differentiable quantization operator.
Due to the rounding function, the operator's gradient is zero almost everywhere, which prevents gradient-based learning. 
To address this, we employ the Straight-Through Estimator (STE) to approximate the gradient:
\begin{equation}
\partial \text{Round}(x)\partial x = 1,
\label{eq:ste}
\end{equation}
where $Round(\cdot)$ is the non-differentiable rounding function. 
Furthermore, following the approach of LSQ~\cite{lsq}, we treat the quantization scale parameter $s$ as a trainable parameter, optimizing it simultaneously with the model weights.

\subsection{Deploy Friendly Constraint}
Our PTQAT framework is designed for practical deployment, ensuring compatibility with existing hardware, operators, and deployment tools. 
We employ a uniform symmetric quantization scheme, which is hardware-friendly and can be integrated directly into matrix multiplication operations without adding computational overhead. 
To demonstrate its real-world applicability, we quantized an 8-bit BEVDepth4D~\cite{BEVDepth} model using our method and successfully deployed it as a TensorRT engine. The results are presented in Section~\ref{subsection:tensorrt}.

\section{Experiment}
\label{sec:exp}
In this section, we evaluate our proposed PTQAT method on various 3D perception tasks, including 3D object detection, BEV segmentation, and occupancy prediction. Unless otherwise specified, our experiments focus on W4-only (4-bit weight) quantization, except for the TensorRT engine deployment, which uses a W8A8 (8-bit weight, 8-bit activation) configuration.

Since quantization for 3D perception tasks remains a relatively underexplored area, we establish performance baselines by implementing standard post-training quantization (PTQ) and quantization-aware training (QAT) methods. 
The results in Section~\ref{subsection:results} demonstrate that our proposed method consistently outperforms these conventional approaches. 
Furthermore, the ablation study in Section~\ref{subsection:ablation} validates our design choices, showing that using mean squared error (MSE) as the difference metric yields superior results compared to other metrics.

\begin{table}
  \centering
  \small
  \resizebox{\columnwidth}{!}{
  \begin{tabular}{@{}c|c|c|c|c|cc@{}}
    \toprule
    Model & Bit-width & Method & Param. & Cost(s) & mAP & NDS \\
    \midrule
     \multirow{4}{*}{BEVDet~\cite{BEVDet}} & \multirow{4}{*}{W4} & / & / & / & 26.3 & 33.7 \\
     & & PTQ Only & / & / & 16.4 & 25.2 \\
     & & QAT Only & 44.3M & 2548 & 27.3 & 35.2 \\
    & & PTQAT & 31.1M & 2316 & \textbf{27.9} & \textbf{35.5} \\ 
    \midrule
     \multirow{4}{*}{BEVDepth4D~\cite{BEVDepth}}& \multirow{4}{*}{W4} & / & / & / & 40.8 & 52.2 \\
    & & PTQ Only & / & / & 35.4 & 47.8 \\
    & & QAT Only & 76.6M & 35635 & 39.7 & 51.3 \\
   & & PTQAT & 26.8M & 32396 & \textbf{40.0} & \textbf{51.6} \\    
    \midrule
    \multirow{4}{*}{BEVFormer~\cite{Bevformer}} & \multirow{4}{*}{W4} & / & / & / & 41.6 & 51.7 \\
    & & PTQ Only  & / & / & 23.2 & 37.6 \\
    & & QAT Only & 69.0M & 47268 & 38.0 & 48.8 \\
    & & PTQAT & 18.7M & 44493 & \textbf{38.6} & \textbf{49.1} \\    
    \midrule
     \multirow{4}{*}{SparseBEV~\cite{SparseBEV}}& \multirow{4}{*}{W4} & / & / & / & 45.4 & 55.5 \\
    & & PTQ Only & / & / & 30.5 & 42.9 \\
    & & QAT Only & 44.6M & 15658 & 43.6 & 54.3 \\
    & & PTQAT & 32.7M & 13849 & \textbf{44.6} & \textbf{55.2} \\    
    \bottomrule
  \end{tabular}
  }
  \caption{Comparison of 3D object detection results on the nuScences val set. The first row of every model is the float accuracy. Param. means trainable parameter numbers when quantization, measured in millions. Cost refers to the training time for QAT, measured in seconds.
  Compared to standard QAT, our approach achieves higher mAP and NDS with lower training costs.}
  \label{tab:object-detection}
\end{table}

\subsection{Implementation Details}
\paragraph{Datasets.} We evaluate our method on the nuScenes~\cite{nuscenes} dataset, a large-scale autonomous driving benchmark comprising 1,000 driving scenes, with 850 designated for training and validation and 150 reserved for testing.
For 3D object detection,  model performance is assessed using the nuScenes Detection Score (NDS), a unified criterion that integrates multiple metrics, including mean Average Precision (mAP) and various error types (ATE, ASE, AOE, AVE, and AAE).
For BEV segmentation and occupancy prediction, we use the mean Intersection over Union (mIoU) as the evaluation metric.  This metric measures the average overlap between the predicted results and the ground-truth labels across all classes.

\paragraph{Experimental Settings.} We fine-tune the 3D perception networks on a single A100 GPU for 1 epoch. The learning rate is set to one-tenth of the pre-training value, and the quantization threshold is configured to $\theta=0.01$. 
We apply 4-bit (W4) uniform symmetric quantization exclusively to the weights. 
This weight-only approach is chosen because it not only reduces the model size effectively but also integrates seamlessly into existing matrix multiplication operations without incurring additional computational overhead.

\paragraph{3D Perception Networks.} 
We evaluate our method on a diverse set of networks across three primary 3D perception tasks: 3D object detection, BEV semantic segmentation, and occupancy prediction.
For 3D object detection, we assess both CNN-based and Transformer-based architectures. The former includes BEVDet~\cite{BEVDet} and BEVDepth4D~\cite{BEVDepth}, which construct dense Bird's-Eye View (BEV) features to predict 3D bounding boxes.
The latter is represented by SparseBEV~\cite{SparseBEV}, a query-based model that leverages learnable object queries within a Transformer structure.
For BEV semantic segmentation, we apply our method to BEVFormer~\cite{Bevformer} , a Transformer-based multi-task model capable of predicting both 3D objects and segmentation maps. We also evaluate on the multi-modal models RCBEVDet++~\cite{RCBEVDet++} and Bevcar~\cite{Bevcar}, which fuse camera images with radar point clouds.
Finally, for occupancy prediction, our evaluation includes the camera-based model BEVStereoOcc~\cite{BEVStereo} and the multi-modal model TEOcc~\cite{TEOcc}.

\begin{table}
  \centering
  \small
  \resizebox{\columnwidth}{!}{ 
  \begin{tabular}{@{}c|c|c|cccc@{}}
    \toprule
    Model & Bit-width & Method & Vehicle & Driv. Area & Lane & mIoU\\
    \midrule
     \multirow{4}{*}{RCBEVDet++~\cite{RCBEVDet++}} & \multirow{4}{*}{W4} & / & 54.7 & 80.4 & 50.0 & 61.7 \\
     & & PTQ only & 45.0 & 65.7 & 35.3 & 48.6 \\
     & & QAT Only & 53.9 & 80.8 & 49.5 & 61.4 \\
     & & PTQAT & \textbf{54.1} & \textbf{81.0} & \textbf{49.7} & \textbf{61.6} \\ 
    \midrule
     \multirow{4}{*}{Bevcar~\cite{Bevcar}}& \multirow{4}{*}{W4} & / & 57.3 & 81.8 & 43.8 & 61.0 \\
     & & PTQ Only & 48.0 & 70.2 & 26.8 & 48.3 \\
     & & QAT Only & 56.0 & 80.4 & 41.8 & 59.4 \\
     & & PTQAT & \textbf{56.6} & \textbf{80.9} & \textbf{42.9} & \textbf{60.1} \\    
    \bottomrule
  \end{tabular}
  }
  \caption{Comparison of BEV semantic segmentation results on the nuScences val set for Vehicle, Drivable Area, and Lane. Our method gets better performance on multi-modal models.}
  \label{tab:semantic-segmentation-1}
\end{table}

\subsection{Results on 3D Perception Tasks}
\label{subsection:results}

\paragraph{3D Object Detection.} 
The 3D object detection results, presented in Table~\ref{tab:object-detection}, compare our method against PTQ-only and QAT-only baselines.
The findings reveal that applying a W4 (4-bit weight) uniform symmetric quantization with the PTQ-only approach results in significant performance degradation. 
On BEVFormer~\cite{Bevformer}, for instance, this leads to a 10.2\% drop in NDS and a 13.8\% drop in mAP.
In contrast, our hybrid PTQAT framework surpasses the QAT-only method, achieving superior accuracy while substantially reducing the number of trainable parameters.
When applied to SparseBEV~\cite{SparseBEV}, our method improves NDS by 0.9\% and mAP by 1.0\%, while requiring 25\% fewer trainable parameter numbers and reducing the fine-tuning time by 1800s.
Similarly, for BEVFormer~\cite{Bevformer}, our approach achieves higher accuracy than QAT-only while using only one-quarter of the trainable parameter numbers. 
Across other models, our method consistently yields an average improvement of 0.3\% in NDS and 0.6\% in mAP over the QAT-only baseline.

\paragraph{BEV Semantic Segmentation.} 
The BEV semantic segmentation results are presented in Table~\ref{tab:semantic-segmentation-1} and Table~\ref{tab:semantic-segmentation-2}. 
Table~\ref{tab:semantic-segmentation-1} focuses on multi-modal models that use both camera images and radar point clouds as input. For these models, our method achieves a 0.2\% improvement in mIoU across all categories while fine-tuning fewer weights compared to the QAT-only baseline.
Table~\ref{tab:semantic-segmentation-2} shows the segmentation performance of the camera-based, multi-task model BEVFormer-S~\cite{Bevformer}, whose detection performance was previously presented in Table~\ref{tab:object-detection}. Our method improves its segmentation results, yielding a 0.4\% to 0.5\% gain in the Boundary and Lane categories and increasing the overall mIoU by 0.3\%.

\begin{table}
  \centering
  \small
  \resizebox{\columnwidth}{!}{  
  \begin{tabular}{@{}c|cccc@{}}
    \toprule
    Metrics & BEVFormer-S~\cite{Bevformer} & PTQ Only & QAT Only & PTQAT \\
    \midrule
    Lane & 48.8 & 33.4 & 47.7 & \textbf{48.1} \\
    Pred. Crossing & 33.9 & 23.8 & 33.5 & \textbf{33.6}  \\
    Boundary & 49.5 & 36.5 & 48.7 & \textbf{49.2} \\
    \midrule
    mIoU & 44.1 & 31.2 & 43.3 & \textbf{43.6} \\    
    
    \bottomrule
  \end{tabular}
  }
  \caption{Comparison of BEV semantic segmentation results of multi-task model on the nuScences val set for Lane, Pred. Crossing and Boundary. We achieve higher results on both object detection and semantic segmentation parts for multi-task models.}
  \label{tab:semantic-segmentation-2}
\end{table}

\paragraph{Occupancy Prediction.} 
Table~\ref{tab:occupancy} presents the results for the occupancy prediction task. 
Our method demonstrates superior performance across different model architectures while fine-tuning fewer weights than the baseline QAT-only approach.
Specifically, we achieve a 0.3\% improvement in mIoU on the camera-based BEVStereoOcc~\cite{BEVStereo} model and a 2.0\% improvement on the multi-modal TEOcc~\cite{TEOcc} model. 
These gains underscore the strong generalizability of our proposed framework. 

\begin{table}
  \centering
  \small
  \begin{tabular}{@{}c|c|cccc@{}}
    \toprule
    Model & Bit-width & Method & mIoU \\
    \midrule
     \multirow{4}{*}{BEVStereoOcc~\cite{BEVStereo}}& \multirow{4}{*}{W4} & / & 37.3 \\
    & & PTQ Only & 22.6 \\
    & & QAT Only & 36.5 \\
    & & PTQAT & \textbf{36.8} \\ 
    \midrule
     \multirow{4}{*}{TEOcc~\cite{TEOcc} }& \multirow{4}{*}{W4} & / & 42.9 \\
     & & PTQ Only & 32.0 \\
   & & QAT Only & 40.5 \\
    & & PTQAT & \textbf{42.5} \\    
    \bottomrule
  \end{tabular}
  \caption{Comparison of occupancy prediction results on the nuScences val set. Both camera-based and multi-modal occupancy prediction networks have all benefited from accuracy improvements with our quantization method. }
  \label{tab:occupancy}
\end{table}

\subsection{Deploy TensorRT Engine}
\label{subsection:tensorrt}
This section demonstrates the practical application of our method. by quantizing the BEVDepth4D~\cite{BEVDepth} model to 8-bit precision and deploying it as a TensorRT engine using NVIDIA's open-source tools. The quantitative results of this experiment are detailed in Table~\ref{tab:tensorrt}. 
\begin{figure*}
  \centering
   \includegraphics[width=0.9\linewidth]{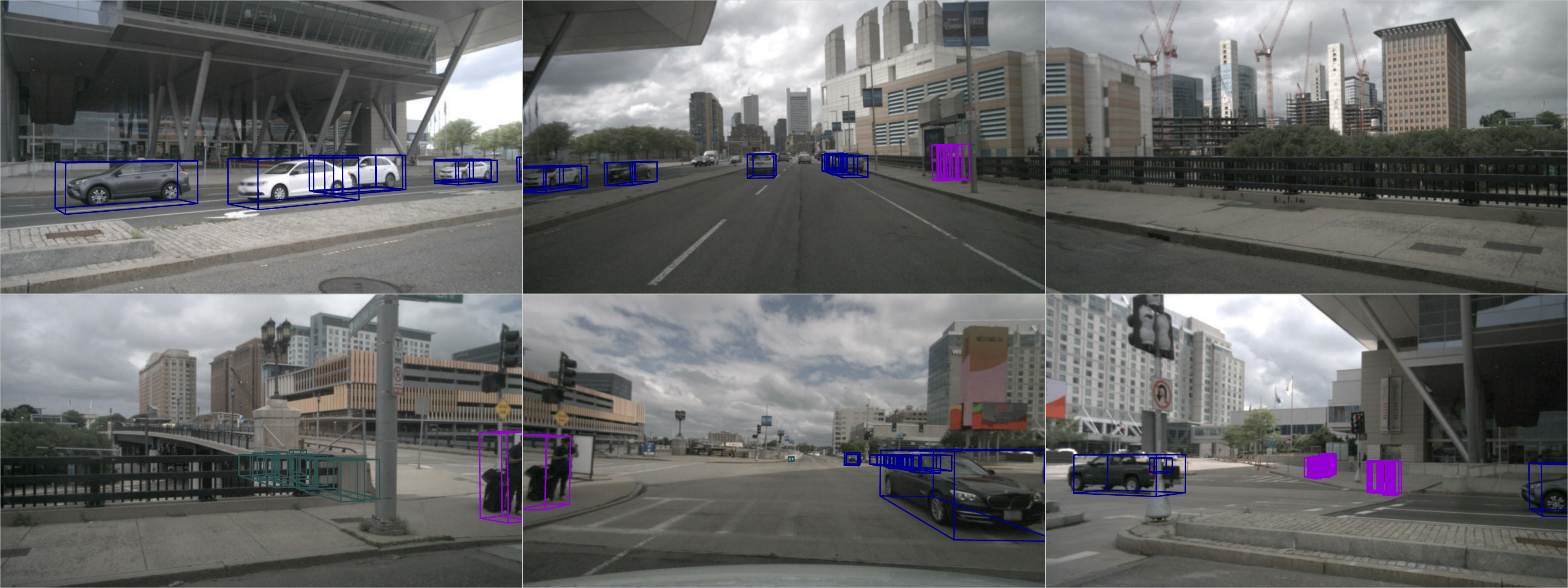}
   \caption{Visualization of 3D object detection results predicted by INT8 quantized BEBDepth4D TensorRT engine. Our method plays a valuable supporting role in streamlining the deployment process from PyTorch models to NVIDIA TensorRT engines. }
   \label{fig:tensorrt_res}
\end{figure*}
While an FP16 TensorRT engine preserves full model accuracy, it suffers from high inference latency and GPU memory consumption. Conversely, applying a standard post-training quantization (PTQ) method to create an 8-bit engine significantly improves efficiency but leads to an unacceptable loss in accuracy.
Our PTQAT framework successfully bridges this gap. It produces an 8-bit TensorRT engine with only a slight performance degradation—a drop of just 2.8\% in mAP and 1.9\% in NDS. This quantized engine achieves an inference speed nearly twice that of its FP16 counterpart while reducing GPU memory usage by over 40\%. The qualitative 3D object detection results from the deployed TensorRT engine are visualized in Figure~\ref{fig:tensorrt_res}.

\begin{table}
  \centering
  \small
  \begin{tabular}{@{}c|ccccc@{}}
    \toprule
    Method & mAP & NDS & FPS & GPU Memory \\
    \midrule
    FP16 & 40.8 & 52.2 & 18 & 860M \\
    NVIDIA PTQ & 32.4 & 44.1 & 35 & 607M \\
    PTQAT (W8A8) & \textbf{38.0} & \textbf{50.3} & 31 & 503M \\
    \bottomrule
  \end{tabular}
  \caption{Performance comparison of different quantization methods on
BEVDepth4D~\cite{BEVDepth} for TensorRT engine. FP16 means keeping weights and inputs as floating-point numbers. W8A8 means quantizing weights and inputs to 8 bits simultaneously. The results indicate that our method can be directly used to help generate a TensorRT engine for practical applications.}
  \label{tab:tensorrt}
\end{table}

\subsection{Ablation Study}
\label{subsection:ablation}
A crucial aspect of our method is the criterion used to identify which layers require fine-tuning.
In this section, we conduct an ablation study to compare several metrics for this purpose: random selection, MSE, cosine similarity, and Huber loss. 
MSE is known for amplifying large errors and its sensitivity to outliers.
Cosine similarity, in contrast, measures only the directional alignment between two vectors, ignoring their magnitude.
Huber loss is a hybrid metric that combines the properties of MSE for small errors and Mean Absolute Error (MAE) for large errors:
\begin{equation}
L_{\delta}=\left\{\begin{array}{ll}
\frac{1}{2}(y-\hat{y})^{2} & \text { if }|y-\hat{y}| \leq \delta \\
\delta|y-\hat{y}|-\frac{1}{2} \delta^{2} & \text { otherwise }
\end{array}\right.
\label{eq:huber}
.
\end{equation}

To ensure a fair comparison, we controlled the number of fine-tuned layers across all tested criteria. For each metric, including a random selection baseline, we adjusted its corresponding hyperparameter $\theta$ to keep the number of unfrozen layers identical.
The results, summarized in Table~\ref{tab:ablation}, demonstrate that MSE achieves the best performance.
The poor performance of cosine similarity is attributable to the fact that while quantization severely alters value magnitudes, it often preserves the overall data distribution. Consequently, the cosine similarity between pre- and post-quantization outputs remains high, making it an ineffective metric for identifying quantization error. In contrast, MSE's effectiveness stems from its tendency to heavily penalize large errors. This sensitivity to outliers aligns with our intuition that these outlier values are a primary cause of performance degradation in quantized models.

\begin{table}
  \centering
  \small
  \begin{tabular}{@{}c|cc@{}}
    \toprule
    Method & mAP & NDS \\
    \midrule
    Huber Loss &  39.6 & 51.4  \\
    Cosine & 39.5 & 51.3 \\
    Random & 39.3 & 51.3 \\  
    MSE-Opposite & 39.3 & 51.1 \\
    MSE & \textbf{40.0} & \textbf{51.6} \\
    \bottomrule
  \end{tabular}
  \caption{Performance comparison of different determination methods on
BEVDepth4D~\cite{BEVDepth}. MSE-Opposite means quantizing layers with large output difference. The results indicate that our MSE method outperforms
other methods.}
  \label{tab:ablation}
\end{table}
\section{Conclusion}

In this paper, we proposed PTQAT, a hybrid quantization framework that synergizes the efficiency of Post-Training Quantization (PTQ) with the performance of Quantization-Aware Training (QAT) for 3D perception networks.
The central principle of our work is the finding that it is more effective to compensate for quantization errors during their propagation path, rather than attempting to correct them only where they originate. Based on this observation, our method employs a layer-wise distortion metric (MSE) to identify layers most impacted by quantization. 

Our framework is designed for practical deployment, using a hardware-friendly uniform symmetric quantization scheme that ensures seamless integration with industry-standard runtimes like TensorRT. The entire process is governed by a single hyperparameter $\theta$ which balances quantization robustness and training efficiency. Extensive experiments on the nuScenes dataset demonstrate that PTQAT is effective and versatile, consistently outperforming standard quantization baselines across diverse 3D perception tasks, including object detection, semantic segmentation, and occupancy prediction.

\section*{Acknowledgments}
This work was supported by National Natural Science Foundation of China (Grant No. 62176007).
{
    \small
    \bibliographystyle{ieeenat_fullname}
    \bibliography{main}
}

\end{document}